\documentclass{article}
\usepackage{spconf,amsmath,graphicx}

\usepackage{caption}
\usepackage{subfigure}
\usepackage{makecell}
\usepackage{multirow}
\usepackage{multicol}
\usepackage{amsmath}
\usepackage{color} 
\usepackage{pifont}
\usepackage{times}  
\usepackage{helvet} 
\usepackage{courier}  
\usepackage[hyphens]{url}  
\usepackage{graphicx} 
\usepackage{caption}
\usepackage{amsthm,amssymb,amsmath,multicol,multirow,threeparttable,booktabs}

\title{A UNIFIED FRONT-END FRAMEWORK FOR ENGLISH TEXT-TO-SPEECH SYNTHESIS}
\name{Zelin Ying$^{\ast\dagger}$\thanks{* Corresponding author. \qquad $\dagger$   Equal contribution.}, Chen Li$^{\dagger}$, Yu Dong, Qiuqiang Kong, Qiao Tian, Yuanyuan Huo, Yuxuan Wang}
\address{\textbf{S}peech, \textbf{A}udio \& \textbf{M}usic \textbf{I}ntelligence (SAMI), ByteDance}
\begin{document}
%
\maketitle
\begin{abstract}
The front-end is a critical component of English text-to-speech (TTS) systems, responsible for extracting linguistic features that are essential for a text-to-speech model to synthesize speech, such as prosodies and phonemes. The English TTS front-end typically consists of a text normalization (TN) module, a prosody word prosody phrase (PWPP) module, and a grapheme-to-phoneme (G2P) module. However, current research on the English TTS front-end focuses solely on individual modules, neglecting the interdependence between them and resulting in sub-optimal performance for each module. Therefore, this paper proposes a unified front-end framework that captures the dependencies among the English TTS front-end modules. Extensive experiments have demonstrated that the proposed method achieves state-of-the-art (SOTA) performance in all modules. 
\end{abstract}
\section{Introduction}
Speech synthesis is a critical technology that has brought significant convenience to the lives of people and has been extensively utilized in various scenarios, including voice assistants and audiobooks \cite{berard2018end}. In English text-to-speech (TTS) synthesis, the front-end plays a vital role by extracting various linguistic features from raw text to provide the acoustic model with sufficient information for synthesizing natural speech.
Enhancing the accuracy of the front-end will lead to improved quality of the TTS synthesized speech.

The English TTS front-end is typically composed of three linguistic-related modules, including: 1) A text normalization (TN) module, which converts non-standard words such as numbers, symbols, and abbreviations into spoken-form words. 2) A prosody word prosody phrase (PWPP) module, which identifies pause boundaries for different duration levels in the sentence. 3) A grapheme-to-phoneme (G2P) module, which converts word sequences into phoneme sequences. For instance, the word ``hello'' is converted to ``HH EH L OW''.

Researchers have focused on developing innovative approaches for each module. For the TN module, Ebden et al. \cite{ebden2015kestrel} proposed a rule-based method using weighted finite-state transducers (WFST) to tokenize the input and convert the classified tokens based on their semiotic class. Due to the limited coverage of rules, Sproat et al. \cite{sproat2017rnn} proposed different recurrent neural network (RNN) architectures to learn the correct normalization function. Zhang et al. \cite{zhang2020hybrid} proposed a hybrid method that combines rules and models to integrate the advantages of a rule-based model and a neural model for text normalization.
For the PWPP module, a simple approach is to add pauses after punctuation marks to indicate prosody. However, recent studies have used sequence tagging neural network models \cite{zheng2018blstm} to predict prosody. Regarding the G2P module, the most common method is to find common words in the lexicon and use neural network models \cite{rao2015grapheme} to predict out-of-vocabulary (OOV) words. Additionally, the part-of-speech (POS) task was proposed to differentiate between homographs, which are written the same but pronounced differently depending on their part-of-speech context.

Although there exist unified front-end frameworks in other languages \cite{pan2020unified,zhang2020unified}, they cannot be compared due to the language-specific differences. In the English TTS front-end, previous studies have mostly focused on improving individual modules, and each component is still usually trained independently \cite{sun2023improving}. However, some issues still persist: 1) No systematic English TTS front-end solution was proposed. 2) Those linguistic-related modules may promote each other, separating them may cause the whole English TTS front-end to be sub-optimal. 3) In the G2P module, some homographs with the same part-of-speech need to be solved.


In this paper, we propose a novel unified front-end framework for English TTS that integrates the TN, PWPP, and G2P modules into a cohesive system to tackle the above challenges. The major contributions of our research are summarized as follows:

\begin{figure*} \vspace{-22pt}
    \centering
    \subfigure[Flowchart of our proposed framework]{
        \includegraphics[scale=0.61]{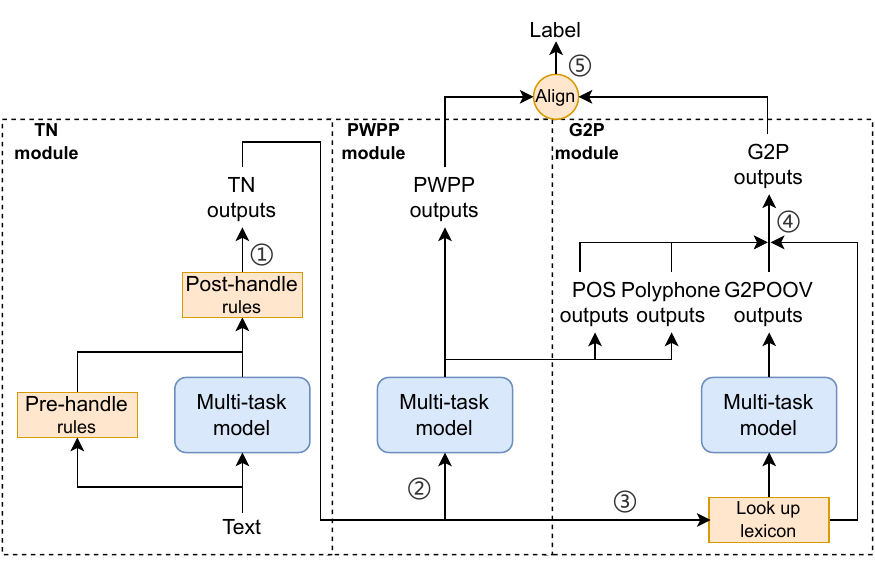}
        \label{fig:flowchart}
    }
    \subfigure[Multi-task model structure]{
        \includegraphics[scale=0.6]{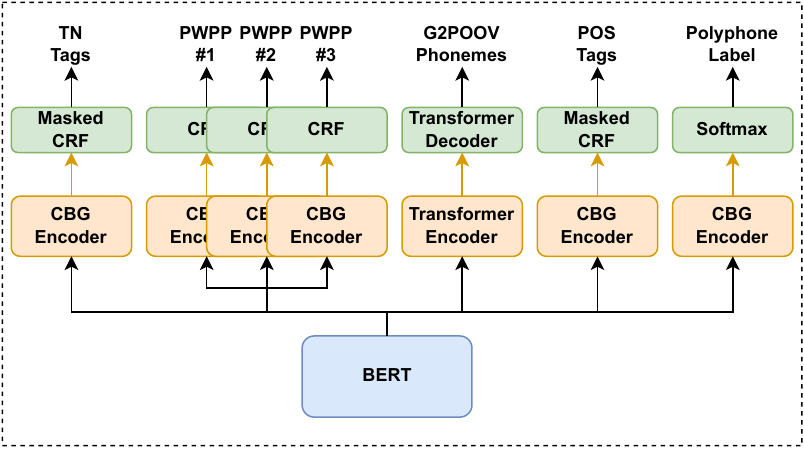}
        \label{fig:model}
    }
    \centering
    \caption{Our unified front-end framework}
    \label{fig:Fig1}
\end{figure*}

\begin{itemize}
\setlength{\itemsep}{0pt}
\setlength{\parsep}{0pt}
\setlength{\parskip}{0pt}
    \item We propose a systematic English TTS front-end framework that employs a shared multi-task model. To the best of our knowledge, this is the first work to unify English TTS front-end tasks.
    \item Our approach exhibits greater flexibility within the TN module, excels at utilizing hierarchical label relationships in the PWPP module, and the introduction of the Polyphone task within the G2P module further enhances the accuracy of homographs with the same part-of-speech.
\end{itemize}

\section{Methodology} \vspace{-10pt}
In this section, we provide an overview of our proposed framework, followed by a detailed description of the workflow for the TN, PWPP, and G2P modules. 

\subsection{Framework Overview}
Figure \ref{fig:flowchart} illustrates the overall flowchart of our proposed framework, which includes the TN, PWPP, and G2P modules. Our framework is composed of a shared multi-task model and well-designed rules. The multi-task model supports five tasks: TN, PWPP, G2POOV, POS, and Polyphone. The TN and PWPP tasks serve their respective modules, while the G2POOV, POS, and Polyphone tasks serve the G2P module. In the TN, PWPP, and G2P modules, the multi-task model is the same, and in the multi-task model, the shared BERT model is jointly trained by all tasks for fine-tuning.

The dataflow in our proposed framework is as follows: \ding{172} We first input raw text to the TN module, which applies a hybrid method combining rules and models to predict the TN outputs; \ding{173} Then, we feed the TN outputs into the shared multi-task model to predict the PWPP, the POS, and the Polyphone outputs; \ding{174} Next, we look up the TN outputs words in the lexicon and directly obtain the pronunciations for the found words, otherwise refer to the G2POOV outputs for the OOV words; \ding{175} Then, we combine phoneme results of the lexicon and OOV words and update homographs pronunciations by the POS and the Polyphone outputs to acquire G2P outputs; \ding{176} Finally, we align PWPP outputs and G2P outputs to obtain the normative label, in which each row shows a phoneme-level feature result.

\subsection{Modules}
\subsubsection{TN Module}
The TN module converts non-standard words, such as numbers, symbols, and abbreviations, into spoken-form words. It includes a multi-task model and an expert-designed rule system. 
Given the input text, we obtain predicted categories using both the model and pre-handle rules separately, and then correct any misclassified categories by the model based on the pre-handle results.
The data for these corrected categories is then fed into post-handle rules to transcribe and obtain the TN transcription results.
For the TN sequence tagging task (TN task) in the multi-task model, we define 19 categories, including ``CARDINAL'', ``DIGIT'', etc. The category ``O'' indicates ``other'' and does not need to be normalized. Each normalized category is accompanied by the beginning, inside, end, and single (BIES) positional tags indicating word boundaries information. We combine the categories, their word boundaries, and the ``O'' category to form $19 \times 4 + 1 = 77$ labels for sequence tagging. The architecture of the TN task model, shown in Figure \ref{fig:model}, consists of a BERT \cite{kenton2019bert} model, a 1-D \textbf{C}onvolution \textbf{B}ank and bidirectional \textbf{G}RU (CBG) \cite{wang2017tacotron} encoder, and a masked \textbf{C}onditional \textbf{R}andom \textbf{F}ields (CRF) \cite{wei2021masked} layer. We use a CRF loss to train the TN task.
Compare to other English TN studies such as \cite{sproat2017rnn} and \cite{sproat2016rnn}, the previous approach relies too heavily on the model and places too much trust in its ability, resulting in difficulties controlling the final output, such as whether to transcribe ``911'' as ``nine hundred and eleven'' or ``nine one one''. The model frequently struggles in these situations. Our hybrid approach, which utilizes both rules and models, effectively handles these issues and is also flexible enough to accurately transcribe hot words.

\subsubsection{PWPP Module}
The PWPP module identifies pause boundaries for different duration levels in sentences, relying on the PWPP sequence tagging task (PWPP task) outputs of the multi-task model. In the PWPP task, pause duration time is divided into three levels, ranging from low to high: prosody word level \#1, prosody phrase level \#2, and intonation phrase level \#3. Instead of directly tagging the three prosody levels, a hierarchical sequence tagging structure is employed to independently tag each prosody level. Specifically, each prosody level is transformed into a binary classification task to predict the probability of the corresponding prosody levels after the words. Finally, the PWPP outputs are determined by adopting the highest predicted prosody level.
Figure \ref{fig:model} shows that the model structure of each prosody level is composed of a BERT model, followed by a CBG encoder and a CRF layer. We use a CRF loss to train the binary classification task of each prosody level and sum all prosody level losses to obtain the PWPP task loss.

\subsubsection{G2P Module}
\begin{figure} 
    \centering
    \includegraphics[scale=0.6]{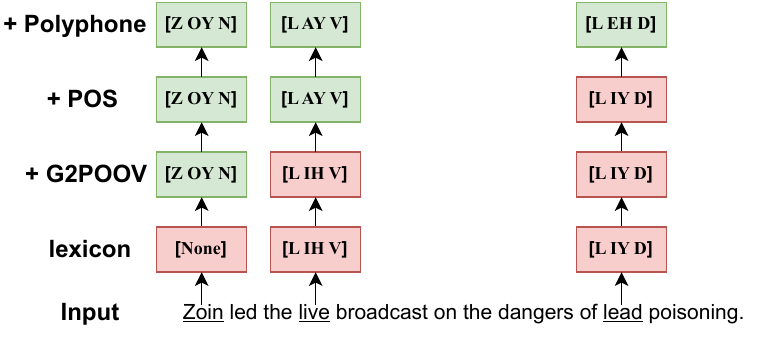}
    \caption{G2P module workflow, red blocks mean wrong phonemes and green blocks mean correct phonemes}
    \label{fig:g2p_module}
\end{figure}

The G2P module converts word sequences into phoneme sequences, which depends on the lexicon and the G2POOV task to achieve candidate pronunciations of each word. Then, we employ the POS and the Polyphone tasks to select the pronunciations of homographs. 
The G2POOV sequence to sequence task (G2POOV task) is used to generate phoneme sequences for the OOV words.
We define 61 characters, including letters, numbers, and punctuation. We expect the model to convert the input characters to output phoneme sequences of 73 phonemes. As shown in Figure \ref{fig:model}, the model architecture of the G2POOV task is composed of a BERT model, followed by a Transformer \cite{vaswani2017attention} encoder and a Transformer decoder. The input to the G2POOV task is an OOV word such as ``Zoin'', which will be tokenized by character level to obtain ``Z o i n'' sequence to output phoneme sequences such as ``Z OY N''. We use a cross entropy loss to train the G2POOV task.
The POS sequence tagging task (POS task) aims to distinguish homographs pronunciations with different part-of-speech. We define 24 part-of-speech categories for the POS task, such as ``Noun'' and ``Verb''. Similar to the TN module, we combine categories, their BIES word boundaries, and the ``O'' category to form $97$ labels for sequence tagging. The model architecture of the POS task is similar to the TN task and is shown in Figure \ref{fig:model}. We use a CRF loss to train the POS task.
The Polyphone sequence classification task (Polyphone task) mainly optimizes homographs with different pronunciations in the same part-of-speech depending on context. We ordered the possible pronunciations of each word as classification labels. For example, the noun ``lead'' pronounced ``L IY D'' for one possible meaning and ``L EH D'' for another. The model architecture of the Polyphone task is composed of a BERT model, followed by a CBG encoder and a softmax classifier, as shown in Figure \ref{fig:model}. The loss for training the Polyphone task is a cross entropy loss. 
To sum up, the workflow of the G2P module is shown in Figure \ref{fig:g2p_module}. Common words will obtain pronunciations from the lexicon, OOV words will obtain pronunciations from the G2POOV task, and homographs pronunciations will be updated based on the POS and Polyphone tasks, to obtain the G2P outputs.



\section{Experimental Evaluations}
\subsection{Experimental Settings}

We conduct extensive experiments in the TN, PWPP, and G2P modules. For evaluating the TN module, we use the Google open source English text normalization dataset \cite{sproat2016rnn} that includes 755,441 sentences. We compare our method with the SOTA method Seq2Edits \cite{stahlberg2020} and other recently proposed methods, such as RNN-based \cite{ro2022transformer} and Transformer-based \cite{zhang2019neural} methods. We use the sentence error rate (SER) as the evaluation metric, which measures the proportion of incorrect sentences to the total sentences. A sentence is considered incorrect if the predicted output does not exactly match the reference.
For the evaluation of the PWPP module, we use an internal dataset including 100,026 sentences. We compare our proposed hierarchical sequence tagging method, which independently predicts each prosody level, against the traditional sequence tagging method, which utilizes the BERT-CBG-CRF structure to directly predict the tags of three prosody levels. We adopt sequence tagging F1-score as the evaluation metric. 
In the G2P module, we train and validate the POS task using 102,164 sentences, and the Polyphone task using 28,097 sentences, while we utilize a dictionary to train and validate the G2POOV task. Our G2POOV approach is compared with the SOTA method r-G2P \cite{zhao2022r}, as well as other recently proposed methods, including Transformer-based G2P \cite{yolchuyeva2019transformer} and encoder-decoder with global attention \cite{toshniwal2016jointly}, to evaluate performance.
In addition, we gather a real-world test set including 2560 sentences to evaluate the overall performance of the complete G2P module. 
Our evaluation metric is the word error rate (WER), where a word is deemed incorrect if the predicted output does not match the reference exactly.

In our framework, we use a pre-trained multilingual BERT \cite{kenton2019bert} model with 12 layers as the language model. We set the hidden unit of the CBG encoder to 256, and the model dimension of the Transformer to 256 with 4 heads in multi-head attention between the Transformer encoder and decoder. Additionally, the model learning rate is set to $5 \times 10^{-5}$, and we use AdamW \cite{loshchilov2018fixing} as the optimizer.


\begin{table}[] \vspace{-3pt}
\centering
\caption{SERs in the TN module test set}
\begin{tabular}{p{5cm}<\centering p{1.5cm}<\centering}
\hline
\textbf{Methods}         & \textbf{SER (\%)}    \\ \hline
RNN-based \cite{zhang2019neural} & 1.80 \\
Transformer-based \cite{ro2022transformer}        & 1.42 \\
Seq2Edits \cite{stahlberg2020}         & 1.36 \\
\textbf{BERT-CBG-CRF (ours)}    & \textbf{1.19} \\ \hline
\end{tabular}
\label{Tab:tn}
\end{table}

\begin{table}[] \vspace{-2pt}
\centering
\caption{Tagging F1-scores in the PWPP module test set}
\begin{tabular}{p{2cm}<{\centering}p{2.2cm}<{\centering}p{2cm}<{\centering}}
\hline
\textbf{Methods}                                                                             & \textbf{Prosody level} & \textbf{F1-score  (\%)} \\ \hline
\multirow{3}{*}{\begin{tabular}[c]{@{}c@{}}Traditional\\ sequence tagging\end{tabular}} & \#1           & 61.15  \\
                                                                                    & \#2           & 56.37  \\
                                                                                    & \#3           & 82.63  \\ \hline
\multirow{3}{*}{\begin{tabular}[c]{@{}c@{}}\textbf{Hierarchical}\\ \textbf{sequence tagging} \\ \textbf{(ours)} \end{tabular}}   & \#1           & \textbf{90.83}  \\
                                                                                    & \#2           & \textbf{57.65}  \\
                                                                                    & \#3           & \textbf{83.36}  \\ \hline
\end{tabular}
\label{Tab:pwpp}
\end{table}

\subsection{Experimental Results and Analysis}
We first evaluate the SER in the TN module, compared with other methods. Table \ref{Tab:tn} shows that the RNN-based model in \cite{zhang2019neural} achieves an SER of 1.80\%, proving RNN-based model cannot achieve good enough results due to the limited encoding ability. Transformer-based model in \cite{ro2022transformer} achieves an SER of 1.42\%, due to the improved encoding ability of the Transformer encoder and the help of the BERT pre-trained language model. The Seq2Edits approach in \cite{stahlberg2020} treats TN as a sequence of edit operations, uses span-level edits to capture compact local representations and achieves an SER of 1.36\%. 
Our proposed method is based on a BERT model fine-tuned for multiple front-end tasks, and a CBG encoder that considers both local and global text features, plus the help of well-designed rules, therefore achieves the best SER of 1.19\%, outperforms the SOTA method (Seq2Edits) by 0.17\% SER. 

Then, we assess the tagging F1-score of different prosody levels in the PWPP module. As presented in Table \ref{Tab:pwpp}, our proposed hierarchical sequence tagging method outperforms the traditional sequence tagging method. More specifically, the hierarchical sequence tagging method shows a significant improvement over the traditional sequence tagging method in the \#1 prosody level and a slight improvement in the \#2 and \#3 prosody levels. It is due to the hierarchical sequence tagging method considers the constraint that high level prosody belongs to the low level prosody, resulting in better overall performance.


Next, we evaluate the G2POOV task and the entire G2P module. Table \ref{Tab:g2poov} presents the results of various methods for the G2POOV task. The vanilla Transformer \cite{yolchuyeva2019transformer} fails to achieve good performance, with a WER of 22.10\%. \cite{toshniwal2016jointly} uses an encoder-decoder architecture with global attention to capture more global information and improve performance of the model, achieving a WER of 21.69\%. \cite{zhao2022r} introduces three methods for controllable noise to improve robustness of the model, achieving a WER of 19.85\%. 
Our proposed method, which benefits from fine-tuning the BERT model using multiple front-end linguistics-related tasks, achieves the best performance WER of 19.42\%.
In the complete G2P module, we evaluate our system on a real-world test set of 2560 sentences. As shown in Table \ref{Tab:g2p}, using a lexicon to search for word pronunciations results in a WER of 3.83\%. One reason for this is that the lexicon may contain incorrect entries. Another reason is that out-of-vocabulary (OOV) words and homographs cannot be handled. To address this, we add the G2POOV task to the G2P module, enabling the system to handle OOV words, which reduce the WER to 3.42\%. Next, we add the POS task to handle homographs that are pronounced differently depending on their part-of-speech, resulting in a WER decrease to 3.17\%. Finally, we add the Polyphone task to handle homographs that are pronounced differently even within the same part-of-speech, resulting in a WER decrease to 3.09\%. 
These results demonstrate the importance of the G2POOV, POS, and Polyphone tasks in the G2P module.
Since then, our proposed framework consistently achieves the best performance in all TN, PWPP, and G2P evaluations, and the effectiveness is strongly demonstrated.

\begin{table}[] \vspace{-3pt}
\centering
\caption{WERs in the G2POOV task on CMUDict test set } 
\begin{tabular}{cc}  
\hline
\textbf{Methods}               & \textbf{WER (\%)} \\ \hline
Transformer 4x4 \cite{yolchuyeva2019transformer}                             & 22.10       \\
Encoder-decoder + global attention \cite{toshniwal2016jointly}                    & 21.69       \\
r-G2P \cite{zhao2022r}              & 19.85       \\
\textbf{BERT-Transformer (ours)}    & \textbf{19.42}       \\ \hline
\end{tabular}
\label{Tab:g2poov}
\end{table}

\begin{table}[] 
\centering
\caption{WERs in the G2P module real-world sentences}
\begin{tabular}{p{6cm}<\centering c}
\hline
\textbf{Methods}\centering                    & \textbf{WER (\%)} \\ \hline
lexicon                             & 3.83       \\
lexicon + G2POOV                    & 3.42       \\
lexicon + G2POOV + POS              & 3.17       \\
\textbf{lexicon + G2POOV + POS + Polyphone}    & \textbf{3.09}       \\ \hline
\end{tabular}
\label{Tab:g2p}
\end{table}

\section{Conclusion}
In this paper, we propose a unified front-end framework for English text-to-speech synthesis to solve complicated front-end tasks, including TN, PWPP, and G2P modules. 
Our method achieves SOTA performance in all modules, with an SER of 1.19\% in the TN module, F1-scores of 90.83\%, 57.65\%, and 83.36\% for the \#1, \#2, and \#3 prosody levels in the PWPP module, a WER of 19.42\% in the G2POOV task, and a WER of 3.09\% in the complete G2P module.
To the best of our knowledge, this is the first work to unify English front-end tasks. Our proposed method may inspire more research work in the literature and cast a major impact on the front-end for English text-to-speech synthesis.

\vfill\pagebreak

\bibliographystyle{IEEEbib}
\bibliography{strings,refs}

\end{document}